\documentclass[runningheads]{llncs}
\usepackage{graphicx}
\usepackage{comment}
\usepackage{amsmath,amssymb} 
\usepackage{color}

\usepackage{comment}
\usepackage{booktabs}

\newcommand{\tabincell}[2]{\begin{tabular}{@{}#1@{}}#2\end{tabular}}  
\usepackage{caption}

\begin{document}
\pagestyle{headings}
\mainmatter
\def\ECCVSubNumber{2390}  

\title{CPGAN: Content-Parsing Generative Adversarial Networks for Text-to-Image Synthesis} 

\titlerunning{CPGAN}
%
\author{Jiadong Liang\inst{1,\dag} \and
Wenjie Pei\inst{2,\dag} \and
Feng Lu\inst{1,3,*}}
\authorrunning{J. Liang, W. Pei et al.}
%
\institute{State Key Lab. of VR Technology and Systems, School of CSE, Beihang University
\mbox{\and Harbin Institute of Technology, Shenzhen \qquad \and Peng Cheng Laboratory, Shenzhen}}

\renewcommand{\thefootnote}{\fnsymbol{footnote}}
\footnotetext[4]{Both authors contributed equally.}
\footnotetext[1]{Corresponding Author: Feng Lu (lufeng@buaa.edu.cn)

\noindent 
This work was supported by National Natural Science Foundation of China (NSFC) under Grant 61972012 and 61732016.}

\maketitle

\begin{abstract}
Typical methods for text-to-image synthesis seek to design effective generative architecture to model the text-to-image mapping directly. It is fairly arduous due to the cross-modality translation. In this paper we circumvent this problem by focusing on parsing the content of both the input text and the synthesized image thoroughly to model the text-to-image consistency in the semantic level. Particularly, we design a memory structure to parse the textual content by exploring semantic correspondence between each word in the vocabulary to its various visual contexts across relevant images during text encoding. Meanwhile, the synthesized image is parsed to learn its semantics in an object-aware manner. Moreover, we customize a conditional discriminator to model the fine-grained correlations between words and image sub-regions to push for the text-image semantic alignment.
Extensive experiments on COCO dataset manifest that our model advances the state-of-the-art performance significantly (from \textbf{35.69} to \textbf{52.73} in Inception Score). 
\end{abstract}

\section{Introduction}
\begin{figure}[t]
\centering
\includegraphics[height=3.5cm]{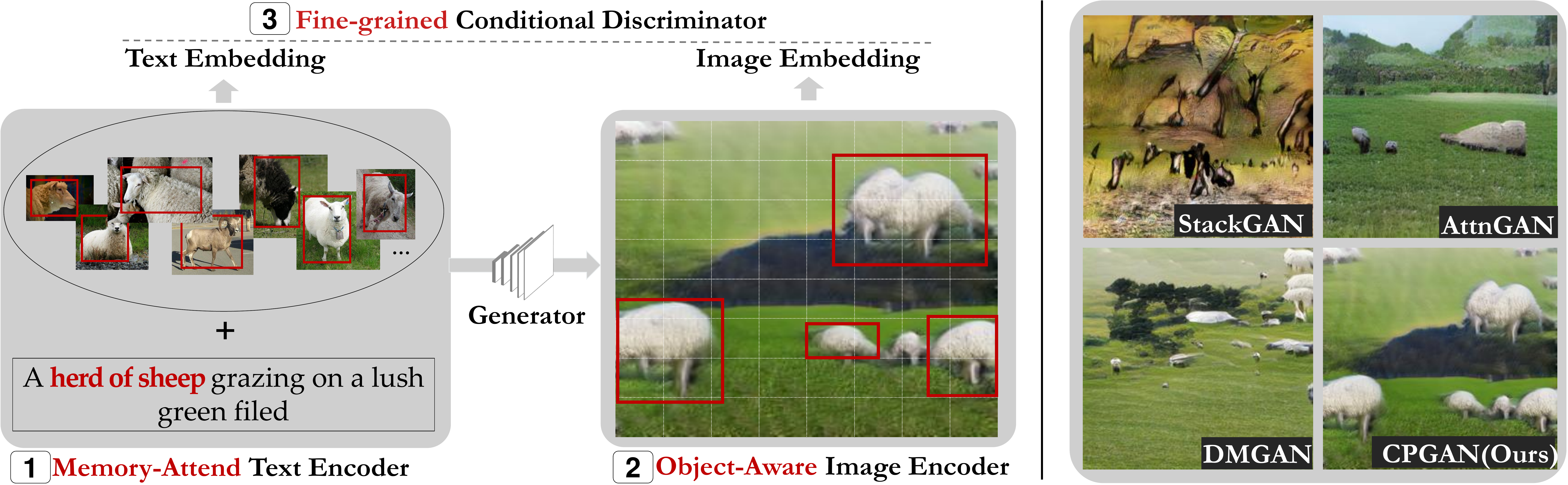}
\caption{Our model parses the input text by a customized memory-attended mechanism and parses the synthesized image in an object-aware manner. Besides, the proposed Fine-grained Conditional Discriminator is designed to push for the text-image alignment in the semantic level.Consequently, our CPGAN is able to generate more realistic and more consistent image than other methods.}
\label{fig:intro}
\end{figure}

Text-to-image synthesis aims to generate an image according to a textual description. The synthesized image is expected to be not only photo-realistic but also consistent with the description in the semantic level. It has various potential applications such as artistic creation and interactive entertainment. Text-to-image synthesis is more challenging than other tasks of conditional image synthesis like label-conditioned synthesis~\cite{odena2016conditional} or image-to-image translation~\cite{isola2017image}. On one hand, the given text contains much more descriptive information than a label, which implies more conditional constraints for image synthesis.
On the other hand, the task involves cross-modality translation which is more complicated than image-to-image translation. Most existing methods~\cite{cha2017adversarial,Hao2017Semantic,hinz2019semantic,hong2018inferring,li2019object,qiao2019mirrorgan,reed2016generative,Hong2019semantics-enhanced-augmented,xu2018attngan,yin2019semantics,zhang2017stackgan,Han2017StackGAN,zhu2019dm-gan:}, for text-to-image synthesis are built upon the GANs~\cite{Goodfellow2014Generative}, which has been validated its effectiveness in various tasks on image synthesis~\cite{Brock2018Large,miyato2018spectral,zhang2018self}. A pivotal example is StackGAN~\cite{zhang2017stackgan} which is proposed to synthesize images iteratively in a coarse-to-fine framework by employing stacked GANs. Subsequently, many follow-up works focus on refining this generative architecture either by introducing the attention mechanism~\cite{xu2018attngan,zhu2019dm-gan:} or modeling an intermediate representation to smoothly bridge the input text and generated image~\cite{hinz2019semantic,hong2018inferring,li2019object}. Whilst substantial progress has been made by these methods, one potential limitation is that these methods seek to model the text-to-image mapping directly during generative process which is fairly arduous for such cross-modality translation. Consider the example in Figure~\ref{fig:intro}, both StackGAN and AttnGAN can hardly correspond the word `sheep' to an intact visual picture for a sheep correctly. It is feasible to model the text-to-image consistency more explicitly in the semantic level, which however requires thorough understanding for both text and image modalities. Nevertheless, little attention is paid by these methods to parsing content semantically for either the input text or the generated image. 
Recently this limitation is investigated by SD-GAN~\cite{yin2019semantics}, which leverages the Siamese structure in the discriminator to learn semantic consistency between two textual descriptions.
However, direct content-oriented parsing in the semantic level for both input text and the generated image is not performed in depth.

\noindent In this paper we focus on parsing the content of both the input text and the synthesized image thoroughly and thereby modeling the semantic correspondence between them. On the side of text modality, we design a memory mechanism to parse the textual content by capturing the various visual context information across relevant images in the training data for each word in the vocabulary. On the side of image modality, we propose to encode the generated image in an object-aware manner to extract the visual semantics. The obtained text embeddings and the image embeddings are then utilized to measure the text-image consistency in the semantic space. Besides, we also design a conditional discriminator to push for the semantic text-image alignment by modeling the fine-grained correlations locally between words and image sub-regions. Thus, a full-spectrum content parsing is performed by the resulting model, which we refer to as Content-Parsing Generative Adversarial Networks (CPGAN), to better align the input text and the generated image semantically and thereby improve the performance of text-to-image synthesis. Going back to the example in Figure~\ref{fig:intro}, our CPGAN successfully translates the textual description `a herd of sheep grazing on a greed field' to a correct visual scene, which is more realistic than the generated results of other methods.
We evaluate the performance of our CPGAN on COCO dataset both quantitatively and qualitatively, demonstrating that CPGAN pushes forward the state-of-the-art performance by a significant step. 
Moreover, the human evaluation performed on a randomly selected subset from COCO test set consistently shows that our model outperforms other two methods(StackGAN and AttnGAN).
To conclude, the idea of our CPGAN to \textbf{parse the content on both the text side (by MATE, in Sec. 3.2) and the image side (by OAIE, in Sec. 3.3)} is novel, which tackles the cross\text{-}modality semantic alignment problem effectively and clearly distinguishes our CPGAN from existing methods. Along with a customized fine-grained conditional discriminator (FGCD, in Sec. 3.4), the CPGAN pushes forward the state-of-the-art performance significantly, from 35.69 to 52.73 in Inception Score.

\section{Related Work}
\smallskip\noindent\textbf{Text-to-Image Synthesis.}
Text-to-image synthesis was initially investigated based on pixelCNN~\cite{reed2017parallel,Reed2017structure}, which suffers from highly computational cost during the inference phase. Meanwhile, the variational autoencoder (VAE)~\cite{mansimov2015generating} was applied to text-to-image synthesis. 
A potential drawback of VAE-based synthesis methods is that the generated images by VAE tend to be blurry presumably. This limitation is largely mitigated by the GANs~\cite{Goodfellow2014Generative}, which was promptly extended to various generative tasks in computer vision~\cite{Brock2018Large,isola2017image,miyato2018spectral,zhang2018self,zhu2017unpaired,niu2019pathological,cao2019makeup,liu2020separate,liu2020unsupervised}.
After Reed~\cite{reed2016generative} made the first attempt to apply GAN to text-to-image synthesis, many follow-up works~\cite{hinz2019semantic,hong2018inferring,li2019object,qiao2019mirrorgan,xu2018attngan,zhang2017stackgan,Han2017StackGAN,zhu2019dm-gan:} focus on improving the generative architecture of GAN to refine the quality of generated images. A well-known example is StackGAN~\cite{zhang2017stackgan,Han2017StackGAN}, which proposes to synthesize images in a coarse-to-fine framework. Following StackGAN, AttnGAN~\cite{xu2018attngan} introduces the attention mechanism which was widely used in computer vision tasks~\cite{yu2019see,lv2019attention} to this framework. DMGAN~\cite{zhu2019dm-gan:} further refines the attention mechanism by utilizing a memory scheme. MirrorGAN~\cite{qiao2019mirrorgan} develops a text-to-image-to-text cycle framework to encourage text-image consistency. Another interesting line of research introduces an intermediate representation as a smooth bridge between the input text and the synthesized image~\cite{hinz2019semantic,hong2018inferring,li2019object,yuan2019bridge}. 
To improve the semantic consistency between the generated image and the input text, ControlGAN~\cite{li2019controllable} applies
the matching scheme of DAMSM in AttnGAN~\cite{xu2018attngan} in all
3-level discriminators. In contrast, our Fine-Grained Conditional Discriminator (FGCD) proposes a novel discriminator structure to capture the local semantic correlations between each caption word and image regions.
Whist these methods have brought about substantial progress, they seek to model the text-to-image mapping directly during generative process. Unlike these methods, we focus on content-oriented parsing of both text and image to obtain a thorough understanding of involved multimodal information. Recently Siamese network is leveraged to explore the semantic consistence either between two textual descriptions by SD-GAN~\cite{yin2019semantics} or two images by SEGAN~\cite{Hong2019semantics-enhanced-augmented}. 
LeicaGAN~\cite{qiao2019learn} adopts text-visual co-embeddings to replace input text with corresponding visual features.
Lao et al.~\cite{lao2019dual} parses the input text by learning two variables that are
disentangled in the latent space.
Text-SeGAN~\cite{cha2019adversarial} focuses on devising a specific discriminator to regress the semantic relevance between text and image. CKD~\cite{yuan2019ckd} parses the image content by a hierarchical semantic representation to enhance the semantic consistency and visual quality of synthesized images. 
However, deep content parsing in the semantic level for both text and image modalities is not performed. 

\smallskip\noindent\textbf{Memory Mechanism.}
Memory networks were first proposed to tackle the limited memory of recurrent networks~\cite{Weston2015,sukhbaatar2015end-to-end}. It was then extensively applied in tasks of natural language processing (NLP)~\cite{das2017question,feng2017memory-augmented,maruf2017document,wang2018target-sensitive} and computer vision (CV)\cite{ma2017visual,mohtarami2018automatic,pei2019memory}. Different from the initial motivation of memory networks that is to enlarge the modeling memory, we design a specific memory mechanism to build the semantic correspondence between a word to all its relevant visual features across training data during text parsing. 

\section{Content-Parsing Generative Adversarial Networks}
\begin{figure}[t]
\centering
\includegraphics[height=3.5cm]{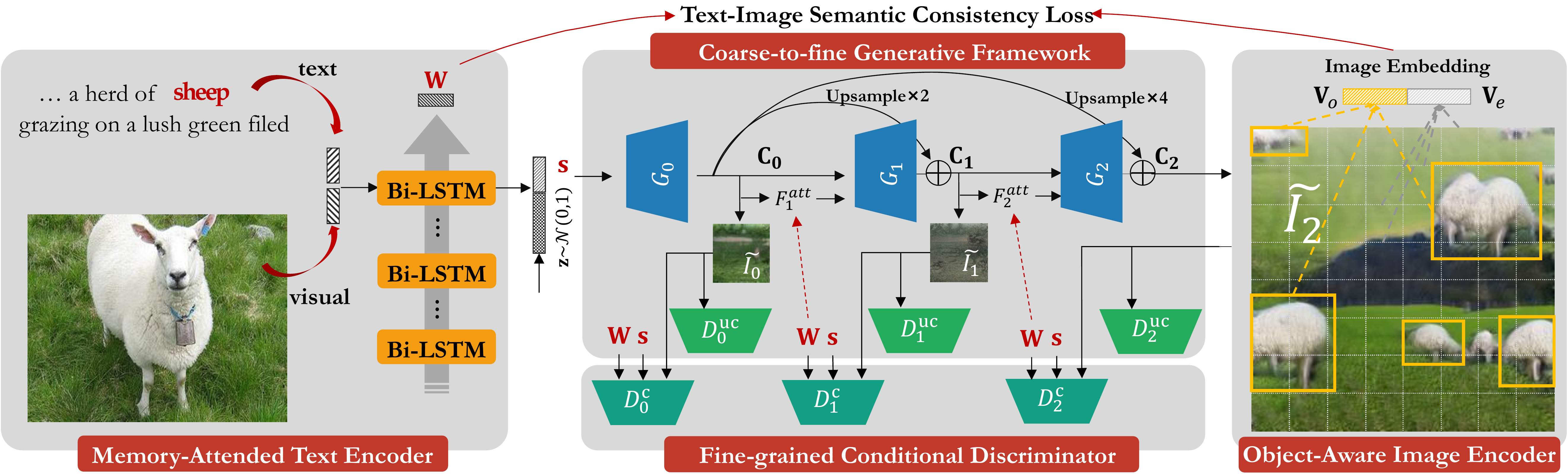}
\caption{Architecture of the proposed CPGAN. It follows the coarse-to-fine generative framework. We customize three components to perform content parsing: Memory-Attended Text Encoder for text, Object-Aware Image Encoder for image, and Fine-graind Conditional Discriminator for the text-image alignment.}
\label{fig:model}
\end{figure}

The proposed Content-Parsing Generative Model for text-to-image synthesis focuses on parsing the involved multimodal information by three customized components. To be specific, the Memory-Attended Text Encoder employs the memory structure to explore the semantic correspondence between a word and its various visual contexts; the Object-Aware Image Encoder is designed to parse the generated image in the semantic level; the Fine-grained Conditional Discriminator is proposed to measure the consistency between the input text and the generated image for guiding optimization of the whole model. 
We will first present the overall architecture of the proposed CPGAN illustrated in Figure~\ref{fig:model}, which follows the coarse-to-fine generative framework, then we will elaborate on the three aforementioned components specifically designed for content parsing. 

\subsection{Coarse-to-fine Generative Framework}
\label{sec:generative_framework}
Our proposed model synthesizes the output image from the given textual description in the classical coarse-to-fine framework, which has been extensively shown to be effective in generative tasks~\cite{li2019object,qiao2019mirrorgan,xu2018attngan,yin2019semantics,zhang2017stackgan,Han2017StackGAN}. As illustrated in Figure~\ref{fig:model}, the input text is parsed by our Memory-Attended Text Encoder and the resulting text embedding is further fed into three cascaded generators to obtain coarse-to-fine synthesized images. Two different types of loss functions are employed to optimize the whole model jointly: 1) Generative Adversarial Losses to push the generated image to be realistic and meanwhile match the descriptive text by training adversarial discriminators and 2) Text-Image Semantic Consistency Loss to encourage the text-image alignment in the semantic level. 

\noindent Formally, given a textual description $X$ containing $T$ words, the parsed text embeddings by the Memory-Attended Text Encoder (Sec.~\ref{sec:text_enc}) are denoted as: $\mathbf{W}, \mathbf{s} = \text{TextEnc}(X)$.
\noindent Herein $\mathbf{W} = \{\mathbf{w}_1, \mathbf{w}_2, \dots \mathbf{w}_T \}$ consists of embeddings of $T$ words in which $\mathbf{w}_t \in \mathbb{R}^{d}$ denotes the embedding for the $t$-th word. $\mathbf{s} \in \mathbb{R}^{d}$ is the global embedding for the whole sentence. Three cascaded generators $\{G_0, G_1, G_2\}$ are then employed to sequentially synthesize coarse-to-fine images $\{\widetilde{I}_0, \widetilde{I}_1, \widetilde{I}_2\}$. We apply similar structure as Generative Network in AttnGAN~\cite{xu2018attngan}:
\begin{align}
\widetilde{I}_0, \mathbf{C_0} = G_0(\mathbf{z}, \mathbf{s}),\qquad
\widetilde{I}_i, \mathbf{C_i} = G_i(\mathbf{C_{i-1}}, F^{att}_i(\mathbf{W}, \mathbf{C_{i-1}})), i=1, 2,
\label{eqn:generation}
\end{align}
\noindent
 where $\mathbf{C}_i$ are the generated intermediate feature maps by $G_i$ and $F^{att}_i$ is an attention model designed to attend to the word embeddings $\mathbf{W}$ to each pixel of $\mathbf{C_{i-1}}$ in $i$-th generation stage. Note that the first-stage generator $G_0$ takes as input the noise vector $\mathbf{z}$ sampled from a standard Gaussian distribution to introduce the randomness. In practice, $F^{att}_i$ and $G_i$ are modeled as convolutional neural networks (CNNs), which are elaborated in the supplementary material. Different from AttnGAN, we introduce extra residual connection from $\mathbf{C_0}$ to $\mathbf{C_1}$ and $\mathbf{C_2}$ (via up-sampling) to ease the information propagation between generators.  

\noindent To optimize the whole model, the generative adversarial losses are utilized
by training generators and the corresponding discriminators alternately.
In particular, we train two discriminators for each generative stage: 1) an unconditional discriminator $D^{\text{uc}}$ to push the synthesized image to be realistic and 2) a conditional discriminator $D^{\text{c}}$ to align the synthesized image and the input text. 
The generators are trained by minimizing following adversarial losses:
\begin{align}
    \mathcal{L}_G = \sum_{i=0}^2 \mathcal{L}_{G_i}, \quad \mathcal{L}_{G_i} = -\frac{1}{2} \mathbb{E}_{\widetilde{I}_i \sim p_{G_i}} D_i^{\text{uc}}(\widetilde{I}_i) -\frac{1}{2} \mathbb{E}_{\widetilde{I}_i \sim p_{G_i}} D_i^{\text{c}}(\widetilde{I}_i, X).
\label{eqn:LG}
\end{align}
\noindent
Accordingly, the adversarial loss for the corresponding discriminators in  the $i$-th generative stage is defined as:
\begin{small}
\begin{align}
    \begin{split}
    \mathcal{L}_{D_i}
    &= \frac{1}{2} \mathbb{E}_{I_i \sim p_{\text{data}_i}} [\max (0, 1-D_i^{\text{uc}}(I_i))] +\frac{1}{3}\mathbb{E}_{\widetilde{I}_i \sim p_{G_i}} [\max (0, 1+ D_i^{\text{uc}}(\widetilde{I}_i) )] \\
     & +\frac{1}{2} \mathbb{E}_{I_i \sim p_{\text{data}_i}} [\max (0, 1-D_i^{\text{c}}(I_i, X))] +\frac{1}{3}\mathbb{E}_{\widetilde{I}_i \sim p_{G_i}} [\max (0, 1+ D_i^{\text{c}}(\widetilde{I}_i, X))]\\
     &+\frac{1}{3} \mathbb{E}_{I_i \sim p_{\text{data}_i}} [\max (0, 1+D_i^{\text{c}}(I_i, \overline{X}))],
    \end{split}
\label{eqn:LD}
\end{align}
\end{small}

\noindent
where $X$ is the input descriptive text and $I_i$ is the corresponding groudtruth image for the $i$-th generative stage. The negative pairs $(I_i, \overline{X})$ are also involved to improve the training robustness. Note that we formulate the adversarial losses in the form of Hinge loss rather than the negative log-likelihood due to the empirical superior performance of Hinge loss~\cite{miyato2018spectral,zhang2018self}. 

\noindent The modeling of unconditional discriminator $D_i^{\text{uc}}$ is straightforward by CNNs (check supplementary material for details), it is however non-trivial to design an effective conditional discriminator $D_i^{\text{c}}$. For this reason, we propose the Fine-grained Conditional Discriminator in Section~\ref{sec:gan_dis}.

\noindent While the adversarial losses in Equation~\ref{eqn:LG},~\ref{eqn:LD} push for the text-image consistency in an adversarial manner by the conditional discriminator, Text-Image Semantic Consistency Loss (TISCL) is proposed to optimize the semantic consistency directly. Specifically, the synthesized image and the input text are encoded respectively, then the obtained image embedding and the text embedding are projected to the same latent space to measure their consistency. We adopt DAMSM~\cite{xu2018attngan} (refer to the supplementary file for details) to compute the non-matching loss between a textual description $X$ and the corresponding image $\widetilde{I}$: 
\begin{align}
    \mathcal{L}_{\text{TISCL}} (\widetilde{I}, X) = \mathcal{L}_{\text{DAMSM}} (\text{ImageEnc}(\widetilde{I}), \text{TextEnc}(X)).
    \label{eqn:tiscl}
\end{align}
\noindent
The key difference between our TISCL and DAMSM lies in encoding mechanisms for both input text (TextEnc) and the synthesized image (ImageEnc). Our proposed Memory-Attended Text Encoder and Object-Aware Image Encoder focus on 1) distilling the underlying semantic information contained in text and image, and 2) capturing the semantic correspondence between them. We will discuss these two encoders in subsequent sections concretely.

\subsection{Memory-Attended Text Encoder}
\label{sec:text_enc}
The Memory-Attended Text Encoder is designed to parse the input text and learn meaningful text embeddings for downstream generators to synthesize realistic images. A potential challenge during text encoding is that a word may have multiple (similar but not identical) visual context information and correspond to more than one relevant images in training data. Typical text encoding methods which encode the text online during training can only focus on the text-image correspondence of the current training pair. Our Memory-Attended Text Encoder aims to capture full semantic correspondence between a word to various visual contexts from all its relevant images across training data. Thus, our model can achieve more comprehensive understanding for each word in the vocabulary and synthesize images of higher quality with more diversity.

\subsubsection{Memory Construction}
\label{sec:memory_cons}
The memory is constructed as a mapping structure, wherein each item maps a word to its visual context representation. To learn the meaningful visual features from each relevant image for a given word, we detect salient regions in each image to the word and extract features from them. There are many ways to achieve this goal. We resort to existing models for image captioning, which is the sibling task of text-to-image synthesis, since we can readily leverage the capability of image-text modeling. In particular, we opt for the Bottom-Up and Top-Down(BUTD) Attention model~\cite{anderson2018bottom} which extracts the salient visual features for each word in a caption at the level of objects. 

\begin{figure}[t]
\centering
\includegraphics[height=4cm]{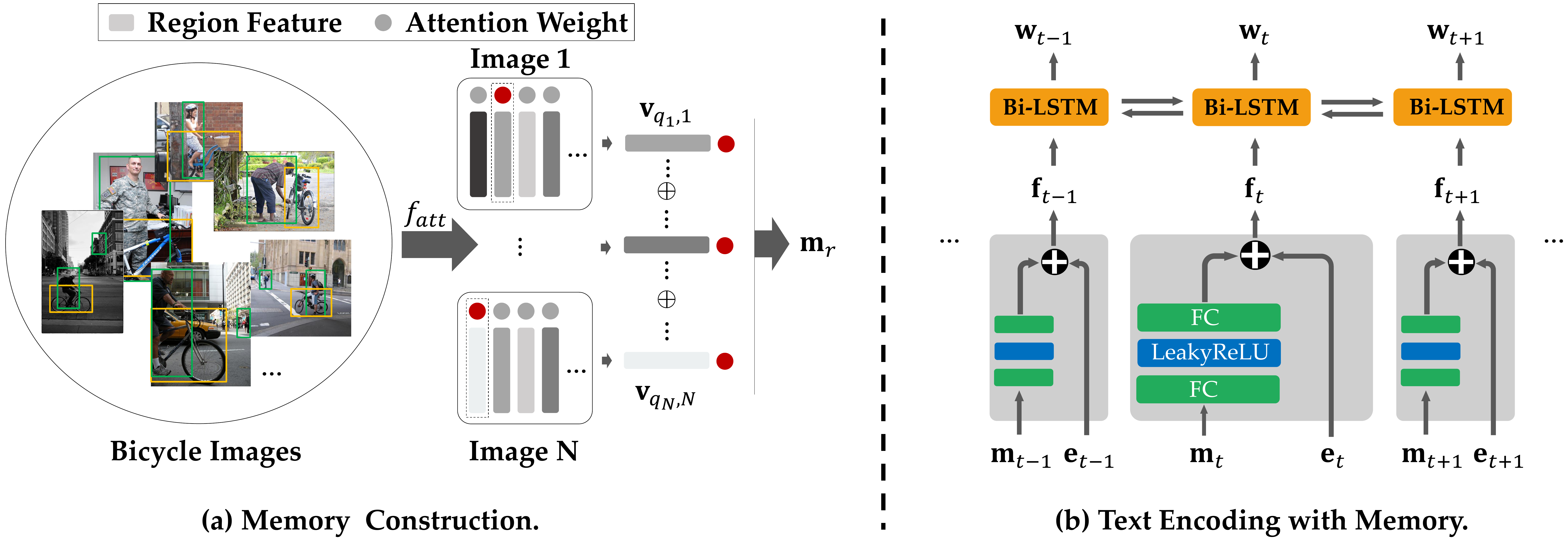}
\caption{(a) The memory $\mathbf{m}_r$ is constructed by considering salient regions from all relevant images across training data. (b) The learned memory and the word embedding are fused via LSTM structure to incorporate temporal information.}
\label{fig:memory}
\end{figure}

\noindent Specifically, given an image-text pair $\langle I, X\rangle$, object detection is first performed on image $I$ by pretrained Yolo-V3~\cite{redmon2018yolov3} to select top-36 sub-regions (indicated by bounding boxes) w.r.t. the confidence score and the extracted features are denoted as $\mathbf{V} = \{\mathbf{v}_1, \mathbf{v}_2, \dots, \mathbf{v}_{36}\}$. Note that we replace the Faster R-CNN with Yolo-V3 for object detection for computational efficiency. Then the pretrained BUTD Attention model is employed to measure the salience of each of 36 sub-regions for each word in the caption (text) $X$ based on attention mechanism. In practice we only retain the visual feature of the most salient sub-region from each relevant image. Since a word may correspond to multiple relevant images, we extract salient visual features for each of the images the word is involved in. As shown in Figure~\ref{fig:memory}\ (a), the visual context features in the memory $\mathbf{m}_r$ for the $r$-th word in the vocabulary is modeled as the weighted average feature: 
\begin{align}
    \begin{split}
        &q_n = \text{argmax}_{i=1}^{36} a_{i, n}, \quad \mathbf{m}_r = \frac{\sum_{n=1}^N a_{q_n, n} \mathbf{v}_{q_n, n}}{\sum_{n=1}^N a_{q_n, n}}, \quad n = 1, \dots, N, 
    \end{split}
\end{align}
\noindent
where $N$ is the number of relevant images in the training data to the $r$-th word; $a_{i, n}$ is the attention weight on $i$-th sub-regions for the $n$-th relevant image and $q_n$ is the index of the most salient sub-region of the $n$-th relevant image. To avoid potential feature pollution, we extract features from top-$K$ most relevant images instead of all $N$ images where $K$ is a hyper-parameter tuned on a validation set.

\noindent The benefits of parsing visual features by such memory mechanism are twofold: 1) extract precise semantic features from the most salient region of relevant images for each word; 2) capture full semantic correspondence between a word to its various visual contexts. 
It is worth mentioning that both Yolo-V3 and BUTD Attention model are pretrained on MSCOCO dataset~\cite{lin2014microsoft} which is also used for text-to-image synthesis, hence we do not utilize extra data in our method. 

\subsubsection{Text Encoding with Memory}
Apart from the learned memory which parses the text from visual context information, we also encode the text by learning latent embedding directly for each word in the vocabulary to characterize the semantic distance among all words. To be specific, we aim to learn an embedding matrix $\mathbf{E} \in \mathbb{R}^{d \times K}$ consisting of $d$-dim embeddings for in total $K$ words in the vocabulary. The learned word embedding $\mathbf{e}_i = \mathbf{E}[:, i]$ for the $i$-th word in the vocabulary is then fused with the learned memory $\mathbf{m}_i$ by concatenation: $\mathbf{f}_i = [\mathbf{e}_i; p(\mathbf{m}_i)]$,
where $p(\mathbf{m}_i)$ is a nonlinear projection function to balance the feature dimensions between $\mathbf{m}_i$ and $\mathbf{e}_i$. In practice, we perform $p$ by two fully-connected layers with a LeakReLU layer~\cite{xu2015empirical} in between, as illustrated in Figure~\ref{fig:memory}\ (b).

\noindent Given a textual description $X$ containing $T$ words, we employ a Bi-LSTM~\cite{huang2015bidirectional} structure to obtain final word embedding for each time step, which incorporates the temporal dependencies between words: $\mathbf{W}, \mathbf{s} = \text{Bi-LSTM}(\mathbf{f}_1, \mathbf{f}_2, ..., \mathbf{f}_T)$.
\noindent
Herein, $\mathbf{W} = \{\mathbf{w}_1, \mathbf{w}_2, \dots \mathbf{w}_T \}$ consists of the embeddings of $T$ words. We use $\mathbf{w}_T$ as the sentence embedding $\mathbf{s}$.

\subsection{Object-Aware Image Encoder}
\label{sec:image_enc}
The Object-Aware Image Encoder is proposed to parse the synthesized image by our generator in the semantic level. The obtained image-encoded features are prepared for the proposed TISCL (Equation~\ref{eqn:tiscl}) to guide the optimization of the whole model by minimizing the semantic discrepancy between the input text and the synthesized image. Thus, the quality of the parsed image features are crucial to the performance of image synthesis by our model.

\noindent Besides learning global features of the whole image, typical way of attending to local image features is to extract features from equally-partitioned image sub-regions~\cite{xu2018attngan}. We propose to parse the image in object level to extract more physically-meaningful features. In particular, we employ Yolo-V3 (pretrained on MSCOCO) to detect salient bounding boxes with top confidence of object detection and learn features from them, which is exactly same as the corresponding operations by Yolo-V3 in the section of memory construction~\ref{sec:memory_cons}. Formally, we extract visual features (1024-dim) of top 36 bounding boxes by Yolo-V3 for a given image $I$, denoted as $\mathbf{V}_o \in \mathbb{R}^{1024 \times 36}$. Another benefit of parsing images in object level is that it is consistent with our Memory-Attended Text Encoder, which parses text based on visual context information in object level.

\noindent The synthesized image in the early stage of training process cannot be sufficiently meaningful for performing object (salience) detection by Yolo-V3, which would adversely affects the image encoding quality. Hence, we also incorporate local features extracted from equally-partitioned sub-regions ($8 \times 8$ in our implementation) like AttnGAN~\cite{xu2018attngan}, which is denoted as $\mathbf{V}_e \in \mathbb{R}^{768 \times 64}$. 
This kind of two-pronged image encoding scheme is illustrated in Figure~\ref{fig:model}.

\noindent Two kinds of extracted features $\mathbf{V}_o$ and $\mathbf{V}_e$ are then projected into latent spaces with the same dimension by linear transformation and concatenated together to derive the final image encoding features $\mathbf{V}_c$:
\begin{align}
\begin{split}
\mathbf{V}'_o = \mathbf{M}_o \mathbf{V}_o + \mathbf{b}_o, \quad \mathbf{V}'_e = \mathbf{M}_e \mathbf{V}_e + \mathbf{b}_e, \quad \mathbf{V}_c = [\mathbf{V}'_o; \mathbf{V}'_e],
\end{split}
\label{equ:Image}
\end{align}
\noindent
where $\mathbf{M}_o \in \mathbb{R}^{256 \times 1024}$ and $\mathbf{M}_e \in \mathbb{R}^{256 \times 768}$ are transformation matrices. The obtained image encoding feature $\mathbf{V}_c \in \mathbb{R}^{256 \times 100}$ is further fed into the DAMSM in Equation~\ref{eqn:tiscl} to compute the TISCL by measuring the maximal semantic consistency between each word of the input text and different sub-region of the image by attention mechanism\footnote{Details are provided in the supplementary file.}.

\subsection{Fine-grained Conditional Discriminator}
\label{sec:gan_dis}

\begin{figure}[t]
\centering
\includegraphics[height=3cm]{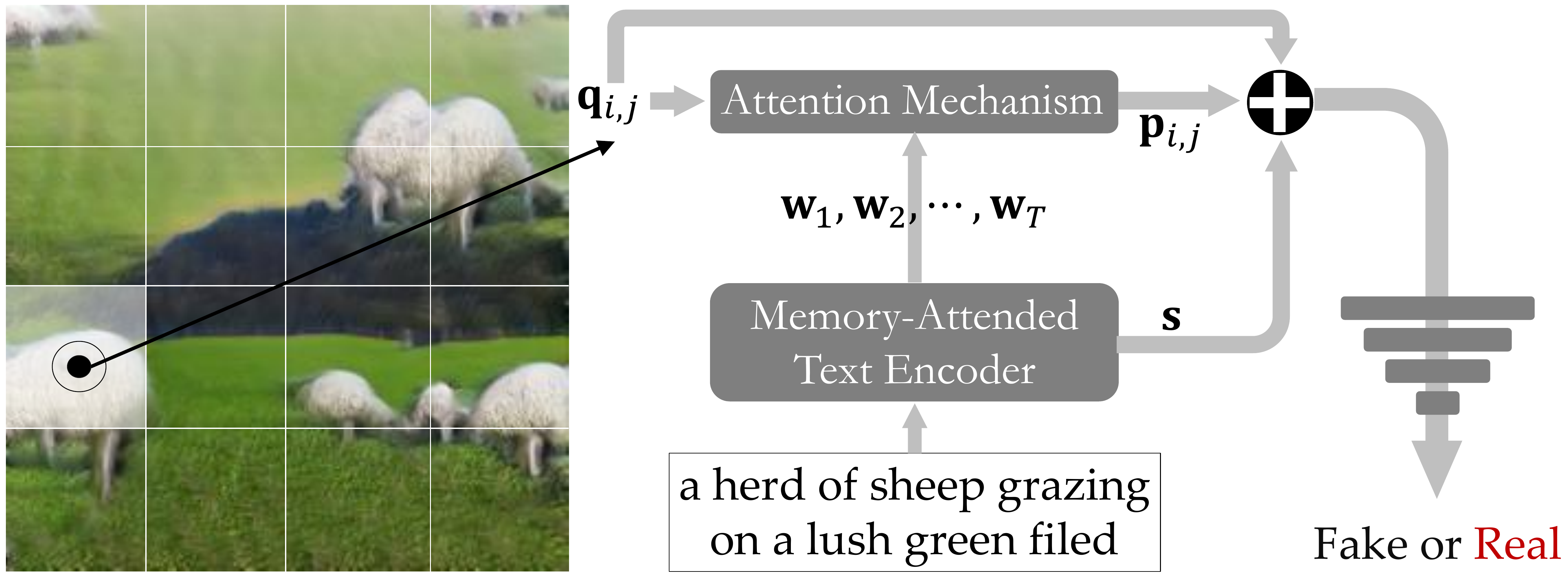}
\caption{The structure of Fine-grained Conditional Discriminator.}
\label{fig:FGCD}
\end{figure}

Conditional discriminator is utilized to distinguish whether a textual caption matches the image in a pair, thus to push the semantic alignment between the synthesized image and the input text by the corresponding adversarial loss. Typical way of designing conditional discriminator is to extract a feature embedding from the text and the image respectively, and then train a discriminator directly on the aggregated features. A potential limitation of such method is that only the global compatibility between the text and the image is considered whereas the local correlations between a word in the text and a sub-region of the image are not explored. Nevertheless, most salient correlations between an image and a caption are always reflected locally. To this end, we propose the Fine-grained Conditional Discriminator, which focuses on modeling local correlations between an image and a caption to measure their compatibility more accurately. 

\noindent Inspired by PatchGAN~\cite{isola2017image}, we partition the image into $N \times N$ patches and extract visual features for each patch. Then learn the contextual features from the text for each patch by attending to each of the word in the text. As illustrated in Figure~\ref{fig:FGCD}, suppose the extracted visual features for the $(i, j)$-th patch in the image are denoted as $\mathbf{q}_{i, j}, i, j \in 1, 2, \dots,N$ and the word features in the text extracted by our text encoder are denoted as $\mathbf{W} = \{\mathbf{w}_1, \mathbf{w}_2, \cdots, \mathbf{w}_T \}$. We compute the contextual features for the $(i, j)$-th patch by attention mechanism:
\begin{small}
\begin{align}
\begin{split}
    &a_n = \frac{\exp({\mathbf{q}_{i,j}^{\top} \mathbf{w}_n})}{\sum_{k=1}^T \exp({\mathbf{q}_{i,j}^{\top} \mathbf{w}_k})}, \quad \mathbf{p}_{i,j} = \sum_{k=1}^{T} a_k \mathbf{w}_k, \quad n=1, 2, \dots, T
\end{split}
\end{align}
\end{small}

\noindent
where $a_n$ is the attention weight for $n$-th word in the text. The obtained contextual feature $\mathbf{p}_{i,j}$ is concatenated together with the visual feature $\mathbf{q}_{i,j}$ as well as the sentence embedding $\mathbf{s}$ for the discrimination to be real for fake. 
Note that the patch size (or the value of $N$) should be tuned to balance between capturing fine-grained local correlations and global text-image correlations.

\section{Experiments}

To evaluate the performance of CPGAN, we conduct experiments on COCO dataset~\cite{lin2014microsoft} which is a widely used benchmark of text-to-image synthesis. 

 \subsection{Experimental Setup}
 \smallskip\noindent\textbf{Dataset.} Following the official 2014 data splits, COCO dataset contains 82,783 images for training and 40,504 images for validation. Each image has 5 corresponding textual descriptions by human annotation. Note that CUB~\cite{wah2011caltech} and Oxford-102~\cite{nilsback2008automated} are not selected since they are too easy (only one object is contained per image) to fully explore the potential of our model. 
 
\smallskip\noindent\textbf{Evaluation Metrics.} We adopt three metrics for quantitative evaluation: Inception score~\cite{salimans2016improved}, R-precision~\cite{xu2018attngan} and SOA~\cite{hinz2019semantic}.
Inception score is extensively used to evaluate the quality of synthesized images taking into account both the authenticity and diversity of images. R-precision is used to measure the semantic consistency between the textual description and the synthesized image.
SOA adopts a pre-trained object detection network to measure whether the objects specifically mentioned in the caption are recognizable in the generated images.
Specifically, it includes two sub-metrics:  SOA\text{-}C (average recall w.r.t. object category) and SOA\text{-}I (average recall w.r.t. image sample), which are defined as:
\begin{small}
\begin{align}
     \mathrm{SOA\text{-}C} = \frac{1}{|C|}\sum_{c\in C}\frac{1}{|I_c|}\sum_{i_c\in I_c}\mathrm{Det}(i_c), \quad \mathrm{SOA\text{-}I} = \frac{1}{\sum_{c\in C}|I_c|}\sum_{c\in C}\sum_{i_c\in I_c}\mathrm{Det}(i_c),
\end{align}
\end{small}

\noindent 
where $C$ and $I_C$ refer to the set of categories and set of images in the category $c$ respectively.
$\mathrm{Det}(i_c) \in \{0, 1\}$ indicates whether the pre-trained detector successfully recognizes an object corresponding to class $c$ in the image $i_c$.

 \smallskip\noindent\textbf{Implementation Details.} Our model is designed based on AttnGAN~\cite{xu2018attngan}, hence AttnGAN is an important baseline to evaluate our model. We make several minor technical improvements over AttnGAN, which yield much performance gain. Specifically, we replace the binary cross-entropy function for adversarial loss with hinge-loss form. Besides, we adopt truncated Gaussian noise~\cite{Brock2018Large} as input noise for synthesis ($\mathbf{z}$ in Equation~\ref{eqn:generation}). We observe that larger batch size in the training process can also lead to better performance. In our implementation, we use batch size of 72 samples instead of 14 samples in AttnGAN. Finally, the hyper-parameters in AttnGAN are carefully tuned. We call the resulting version based on these improvements as AttnGAN$^+$.

\subsection{Ablation Study}
\begin{figure}[t]
\centering
\includegraphics[height=4cm]{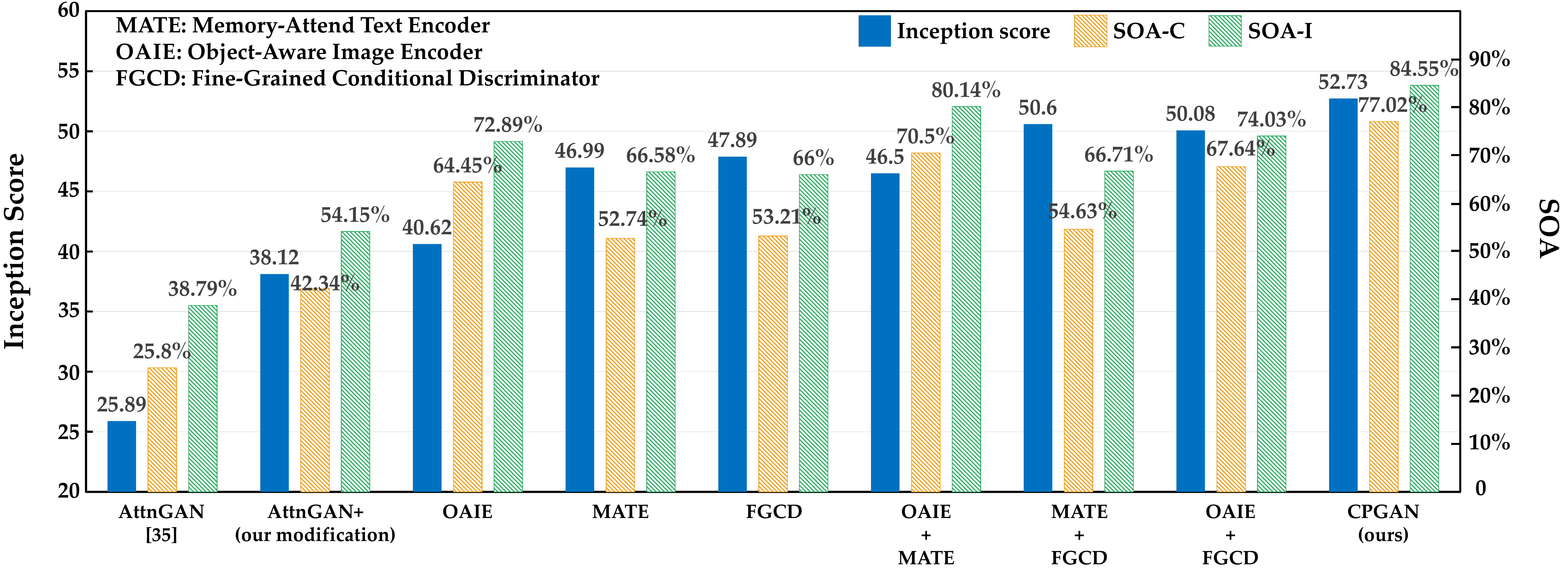}
\caption{Performance of ablation study both in Inception score and SOA.}
\label{fig:ablation}
\end{figure}

 We first conduct experiments to investigate the effectiveness of our proposed three modules respectively, i.e., Memory-Attended Text Encoder (MATE), Object-Aware Image Encoder (OAIE) and Fine-Grained Conditional Discriminator (FGCD). To this end, we perform ablation experiments which begins with AttnGAN$^+$, and then incrementally augments the text-to-image synthesis system with three modules. Figure~\ref{fig:ablation} presents the performance measured by Inception score and SOA of all ablation experiments.
 
 \smallskip\noindent\textbf{AttnGAN$^+$ versus AttnGAN.} It is shown in Figure~\ref{fig:ablation} that AttnGAN$^+$ performs much better than original AttnGAN, which benefits from the aforementioned technical improvements. We observe that increasing the batch size (from 14 to 72) during training brings about the largest performance gain (around 7 points in Inception score). Additionally, fine-tuning the hyper-parameters  contributes another $4$ points of improvement in Inception score to the performance. 
 Besides, the substantial performance gains in SOA show that AttnGAN$^+$ could synthesis images containing more recognizable objects than AttnGAN.
 
 \noindent\textbf{Effect of single module.}
 Equipping the system with each of three proposed modules individually boosts the performance substantially. Compared to AttnGAN$^+$, the performance is improved by 8.9 points, 2.5 points, and 9.8 points by MATE, OAIE and FGCD respectively in Inception score. SOA evaluation results also show large improvements by each of three modules. It is worth noting that OAIE performs best among three modules on SOA metrics emphasizing more on object-level semantics in synthesized images, which in turn validates that OAIE could effectively parse the image in object level. These improvements demonstrate the effectiveness of all three modules. Whilst sharing same generators with AttnGAN$^+$, all our three modules focus on parsing the content of the input text or the synthesized image. Therefore, it is implied that deeper semantic content parsing for the text by the memory-based mechanism helps the downstream generators to understand the input text more precisely. On the other hand, our OAIE encourages generators to generate more consistent images with the input text in object level under the guidance of our TISCL. Besides, FGCD steers the optimization of generators to achieve better alignment between the text and the image by the corresponding adversarial losses. 
 
\noindent\textbf{Effect of combined modules.}
 We then combine every two of three modules together to further augment the text-to-image synthesis system. The experimental results in Figure~\ref{fig:ablation} indicate that the performances in Inception score are further enhanced compared to the results of single-module cases with the exception of MATE + OAIE. We surmise that this is because MATE performs similar operations as OAIE when learning the visual context information from images in the object level for each word in the vocabulary. Nevertheless, OAIE still advances the performances after being mounted over the single FGCD or MATE + FGCD. It can be observed that combined modules also perform much better than the corresponding single module on SOA metrics.
 Employing all three modules leads to our full CPGAN model and achieves the best performance in all metrics, which is better than all other single-module or double-module cases. 

\begin{figure}[!t]
\centering
\includegraphics[height=2.9cm]{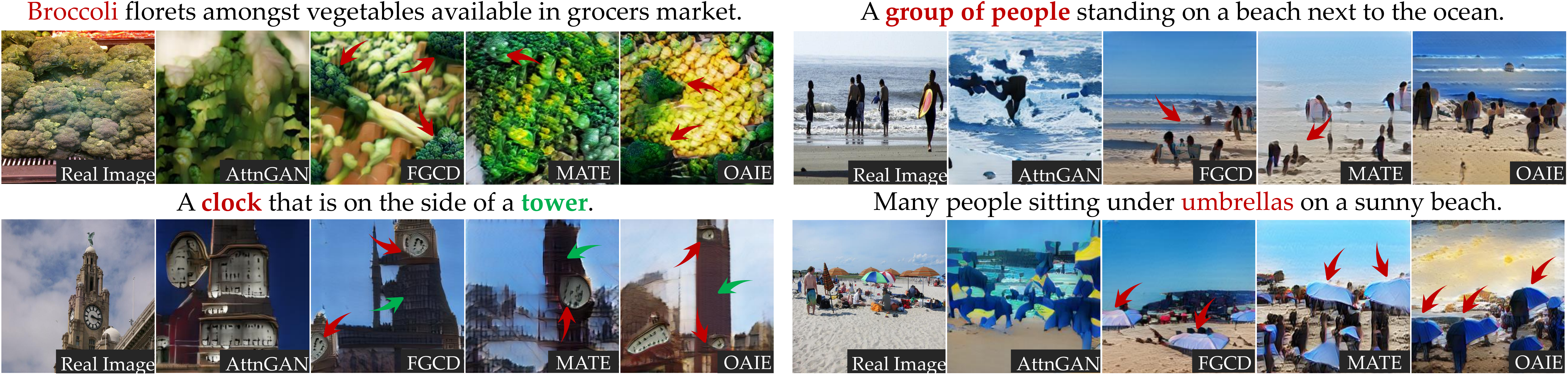}
\caption{Qualitative comparison between different modules of our model for ablation study, the results of AttnGAN are also provided for reference.}
\label{fig:ablation_vis}
\end{figure}

 \noindent\textbf{Qualitative evaluation.}
 To gain more insight into effectiveness of our three modules, we visualize the synthesized images for several examples by systems equipped with different modules and the baseline AttnGAN. Figure~\ref{fig:ablation_vis} presents the qualitative comparison. 
 Compared to AttnGAN, the synthesized images by each of our three modules are more realistic and more consistent with the input text, which again reveals advantages of our proposed modules over AttnGAN. Benefiting from the content-oriented parsing mechanisms, our modules tend to generate more intact and realistic objects corresponding to the meaningful words in the input text, which are indicated with red or green arrows.
 
\subsection{Comparison with State-of-the-arts}
In this set of experiments, we compare our model with the state-of-the-art methods for text-to-image synthesis on COCO dataset. 

\setlength{\tabcolsep}{4pt}
\begin{table}[t]
\begin{center}
\caption{Performance of different text-to-image synthesis models on COCO dataset in terms of Inception score, R-precision SOA-C, SOA-I and model size.}
\label{table:all}
\resizebox{0.9\linewidth}{!}{
\begin{tabular}{l|cccc|c}
\toprule\noalign{\smallskip}
Model & Inception score & R-precision & SOA-C & SOA-I & $\#$Parameters\\
\noalign{\smallskip}
\midrule
\noalign{\smallskip}
Reed~\cite{reed2016generative} & 7.88 $\pm$ 0.07 & $-$ & $-$ & $-$ & $-$ \\
StackGAN~\cite{zhang2017stackgan} &  8.45 $\pm$ 0.03 & $-$& $-$ & $-$ & 996M \\
StackGAN$++$~\cite{Han2017StackGAN} &  8.30 $\pm$ 0.03 & $-$ & $-$ & $-$ & 466M \\
Lao~\cite{lao2019dual} & 8.94 $\pm$ 0.20 & $-$ & $-$ & $-$ & $-$\\
Infer~\cite{hong2018inferring} & 11.46 $\pm$ 0.09 & $-$ & $-$ & $-$ & $-$\\
MirrorGAN~\cite{qiao2019mirrorgan} & 26.47 $\pm$ 0.41 & $-$ & $-$ & $-$ & $-$\\
SEGAN~\cite{Hong2019semantics-enhanced-augmented} & 27.86 $\pm$ 0.31 & $-$ & $-$ & $-$ & $-$\\
ControlGAN~\cite{li2019controllable} & 24.06 $\pm$ 0.60 & $-$ & $-$ & $-$ & $-$\\
SD-GAN~\cite{yin2019semantics} & 35.69 $\pm$ 0.50 & $-$ & $-$ & $-$ & $-$\\
DMGAN~\cite{zhu2019dm-gan:} & 30.49 $\pm$ 0.57 & 88.56\% & 33.44\% & 48.03\% & 223M\\
AttnGAN~\cite{xu2018attngan} & 25.89 $\pm$ 0.47 & 82.98\% & 25.8\% & 38.79\% & 956M\\
objGAN\cite{li2019object} & 30.29 $\pm$ 0.33 &91.05\% &27.14\% &41.24\% & $-$\\
OP\text{-}GAN\cite{hinz2019semantic} & 28.57 $\pm$ 0.17 & 87.90\% &33.11\% &47.95\% & 1019M\\
\midrule
\tabincell{l}{AttnGAN$^+$ \\ \begin{scriptsize}(our modification)\end{scriptsize}}  & 38.12 $\pm$ 0.68 & 92.58\% & 42.34\% &  54.15\% & 956M \\
CPGAN (ours) &\textbf{52.73} $\pm$ 0.61 & \textbf{93.59\%} & \textbf{77.02\%} & \textbf{84.55\%} & 318M\\
\bottomrule
\end{tabular}}
\end{center}
\end{table}
\setlength{\tabcolsep}{1.4pt} 
\smallskip\noindent\textbf{Quantitative Evaluation.} Table~\ref{table:all} reports the quantitative experimental results. Our model achieves the best performance in all four metrics and outperforms other methods significantly in terms of Inception score and SOA, which is owing to joint contributions from all three modules we proposed. Particularly, our CPGAN boosts the state-of-the-art performance by 47.74\% in inception score, 130.32\% in SOA\text{-}C and 78.12\% in SOA\text{-}I.
It proves that the synthesized images by our model not only have higher authenticity and diversity, but also are semantically consistent with the corresponding captions in object level. 
It is worth mentioning that our CPGAN contains much less parameters than StackGAN and AttnGAN, which also follow the coarse-to-fine generative framework. The reduction of model size mainly benefits from two aspects: 1) a negligible amount of parameters are introduced by our proposed MATE and OAIE, 2) The parameter number of three-level discriminators are substantially reduced due to the adoption of Patch-based discriminating behavior in our proposed FGCD.

\begin{figure}[t]
\begin{minipage}[b]{.5\linewidth}
\centering
\includegraphics[width=0.8\linewidth]{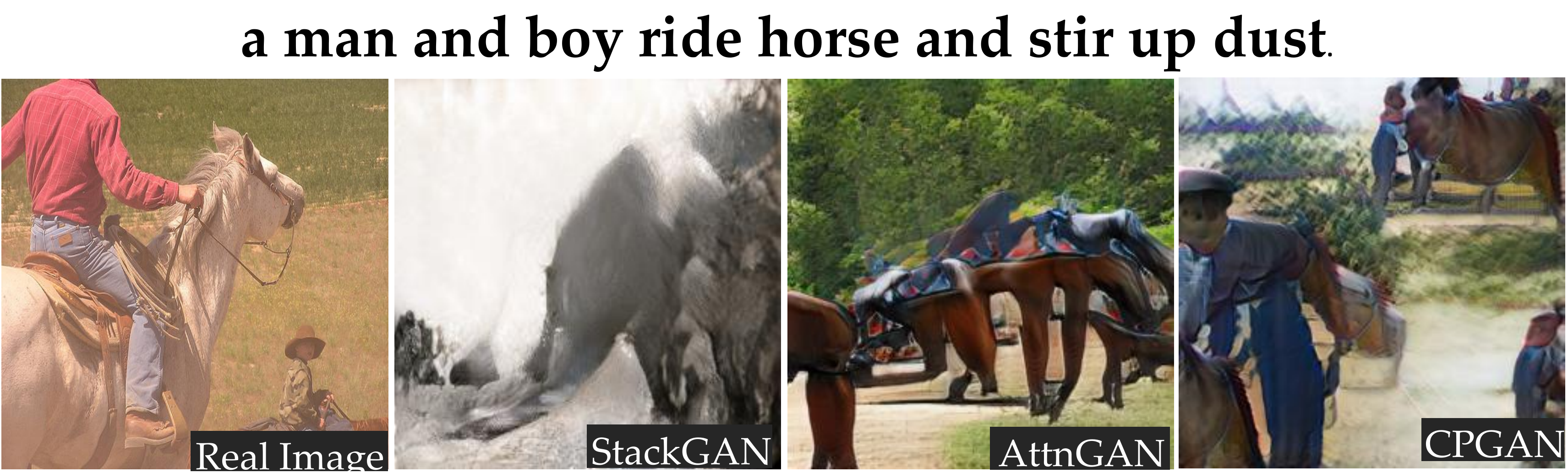}
\caption{Challenging examples. 
}
\label{fig:negative_sample}
\end{minipage}%
\quad \ \ 
\begin{minipage}[b]{.45\linewidth}
\begin{scriptsize}
\centering
\begin{tabular}{lc}
\hline\noalign{\smallskip}
Model & Rank-1 ratio\\
\noalign{\smallskip}
\hline
\noalign{\smallskip}
StackGAN~\cite{zhang2017stackgan} &  7.94\% \\
AttnGAN~\cite{xu2018attngan} & 28.33\%\\
CPGAN(ours)& \textbf{63.73\%}\\
\hline
\end{tabular}
\end{scriptsize}
\captionof{table}{Human evaluation results. 
}
\label{table:human}
\end{minipage}
\end{figure}

\smallskip\noindent\textbf{Human Evaluation.}
As a complement to the standard evaluation metrics, we also perform a human evaluation to
compare our model with two classical models: StackGAN and AttnGAN. We randomly select 50 test samples and ask 100 human subjects to compare the quality of synthesized images by these three models and vote for the best for each sample. Note that three models' synthesized results are presented to human subjects randomly for each test sample. We calculate the rank-1 ratio for each model as the comparison metric, presented in Table~\ref{table:human}. Averagely, our model achieves 63.73\% of votes while AttnGAN wins on 28.33\% votes and StackGAN performs worst. This human evaluation result is consistent with the quantitative results in terms of Inception score in Table~\ref{table:all}. 

\begin{figure}[t]
\centering
\includegraphics[height=4.35cm]{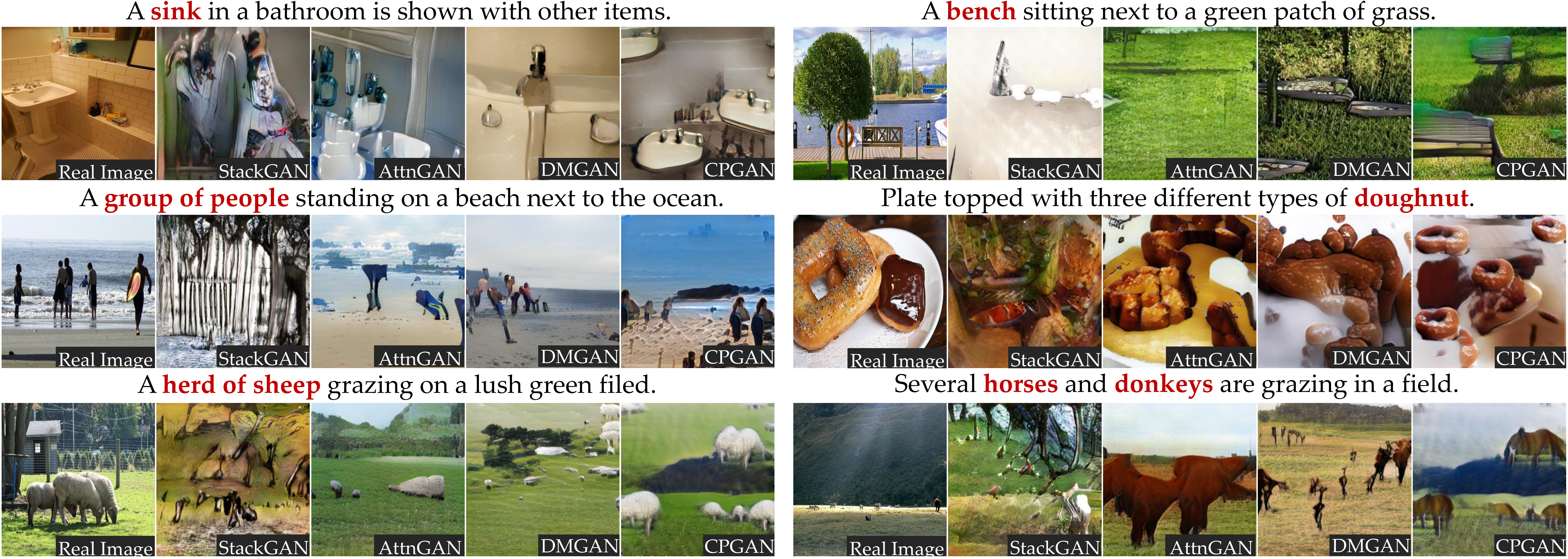}
\caption{Qualitative comparison between our CPGAN with other classical models.}
\label{fig:compare_vis}
\end{figure}

\smallskip\noindent\textbf{Qualitative Evaluation.}
To obtain a qualitative comparison, we visualize the synthesized images on randomly selected text samples by our models and other three classical models: StackGAN, AttnGAN and DMGAN, which is shown in Figure~\ref{fig:compare_vis}. It can be observed that our model is able to generate more realistic images than other two models, like `sheep' , `doughnuts' or `sink'. Besides, the scenes in the generated image by our model are also more consistent with the given text than the other models, such as `bench next to a patch of grass'.

\noindent Image synthesis from text is indeed a fairly challenging task that is far from solved. 
Take Figure~\ref{fig:negative_sample} as a challenging example, all models can hardly precisely interpret the the interaction (`ride') between `man' and `horse'. Nevertheless, our model still synthesizes more reasonable images than other two methods.


\section{Conclusions}

In this work, we have presented the Content-Parsing Generative Adversarial Networks (CPGAN) for text-to-image synthesis. The proposed CPGAN focuses on content-oriented parsing on both the input text and the synthesized image to learn the text-image consistency in the semantic level. Further, we also design a fine-grained conditional discriminator to model the local correlations between words and image sub-regions to push for the text-image alignment. Our model significantly improves the state-of-the-art performance on COCO dataset.

%
%
\bibliographystyle{splncs04}
\bibliography{egbib}
\end{document}